\documentclass[sigconf,authorversion]{acmart}
\AtBeginDocument{%
  }

\setcopyright{acmlicensed}
\copyrightyear{2018}
\acmYear{2018}
\acmDOI{XXXXXXX.XXXXXXX}
\acmConference[Conference acronym 'XX]{Make sure to enter the correct
  conference title from your rights confirmation email}{June 03--05,
  2018}{Woodstock, NY}
\usepackage{soul}
\usepackage{url}
\usepackage{subfigure}
\usepackage[utf8]{inputenc}
\usepackage{algorithm}
\usepackage{algorithmic}
\usepackage{mathtools}
\usepackage{multirow}
\usepackage{stmaryrd}
\usepackage[acronym]{glossaries}
\usepackage{siunitx}
\usepackage[capitalise,sort&compress]{cleveref}
\usepackage[table]{xcolor}
\usepackage{subcaption}
\usepackage{algorithm}

\newcommand{\val}[2]{#1{\tiny{$\pm$#2}}}
\newcommand{\valb}[2]{$\mathbf{#1} {\scriptstyle \, \pm \, #2}$}

\newcommand{\cov}{\ensuremath{K}}  % K or S or \Theta

\newacronym[plural=NNs]{nn}{NN}{Neural Network}
\newacronym[plural=NNs]{mhsa}{MHSA}{Multi-Head Self-Attention}

\begin{document}
\title{Continual Knowledge Consolidation LoRA for Domain Incremental Learning}

\author{Naeem Paeedeh}
\email{naeem.paeedeh@adelaide.edu.au}

\affiliation{
  \institution{School of Computer Science and IT, Adelaide University}
  \city{Adelaide}
  \country{Australia}
}

\author{Mahardhika Pratama}
\email{dhika.pratama@adelaide.edu.au}

\affiliation{
  \institution{School of Computer Science and IT, Adelaide University}
  \city{Adelaide}
  \country{Australia}
}

\author{Weiping Ding}
\email{dwp9988@163.com}

\affiliation{
    \institution{School of Artificial Intelligence and Computer Science, Nantong University}
    \city{Nantong}
    \country{China}
}

\author{Jimmy Cao}
\email{jimmy.cao@adelaide.edu.au}

\affiliation{
  \institution{School of Computer Science and IT, Adelaide University}
  \city{Adelaide}
  \country{Australia}
}

\author{Wolfgang Mayer}
\email{wolfgang.mayer@adelaide.edu.au}

\affiliation{
  \institution{School of Computer Science and IT, Adelaide University}
  \city{Adelaide}
  \country{Australia}
}

\author{Ryszard Kowalczyk}
\email{ryszard.kowalczyk@adelaide.edu.au}

\affiliation{
  \institution{School of Computer Science and IT, Adelaide University}
  \city{Adelaide}
  \country{Australia}
}

\author{Ary Shiddiqi}
\email{ary.shiddiqi@its.ac.id}

\affiliation{
  \institution{Department of Informatics, Institut Teknologi Sepuluh Nopember}
  \city{Surabaya}
  \country{Indonesia}
}

\begin{abstract}
Domain Incremental Learning (DIL) is a sub-branch of continual learning that aims to address the never-ending arrival of new domains without catastrophic forgetting problems. Despite the advent of parameter-efficient fine-tuning (PEFT) approaches, prior works create task-specific LoRAs that overlook shared knowledge across tasks. Inaccurate selection of task-specific LORAs during inference leads to significant drops in accuracy, while existing works rely on linear or prototype-based classifiers, which have suboptimal generalization powers. Our paper proposes continual knowledge consolidation low-rank adaptation (CONEC-LoRA) addressing the DIL problems. CONEC-LoRA is developed from consolidations between task-shared LORA to extract common knowledge and task-specific LORA to incorporate domain-specific knowledge. Unlike existing approaches, CONEC-LoRA integrates the concept of a stochastic classifier whose parameters are sampled from a distribution, thus enhancing the likelihood of correct classifications. Last but not least, an auxiliary network is deployed to optimally predict the task-specific LoRAs for inference and implements the concept of a different-depth network structure in which every layer is connected with a local classifier to leverage intermediate representations. This module integrates the ball-generator loss and transformation module to address the synthetic sample bias problem. Our rigorous experiments demonstrate the advantage of CONEC-LoRA over prior arts in $4$ popular benchmark problems with over $5\%$ margins. 
\end{abstract}

\maketitle

% Note that keywords are not normally used for peer review papers.
\keywords{Continual Learning, Incremental Learning, Domain Incremental Learning}

\section{Introduction}

Continual learning (CL) constitutes a research area of growing interests where the main goal is to develop a learning agent that can accumulate knowledge overtime \cite{chen2018lifelong,Parisi2018ContinualLL,Masana2020ClassIncrementalLS,Zhou2023ClassIncrementalLA}. This task is challenging for a deep neural network (DNN) because of the catastrophic forgetting (CF) problem \cite{Kim2023OnTS} where learning a new task over-writes previously valid parameters thus suffering from significant performance drops on previous tasks. In other words, a plastic model is capable of mastering new tasks, but its performance on previous tasks is compromised. In contrast, a stable model retains its old knowledge but fails to learn new tasks. Hence, the key is to balance the issue of plasticity and stability \cite{Kim2023OnTS}.

There exist three classical approaches in combating the CF problem: regularization-based approach \cite{Kirkpatrick2016OvercomingCF,Mao2021ContinualLV,Paik2019OvercomingCF,Zenke2017ContinualLT,aljundi2018memory,Li2019LearnTG,Schwarz2018ProgressC,Cha2020CPRCR}, memory-based approach \cite{Chaudhry2018EfficientLL,Rebuffi2016iCaRLIC,Chaudhry2019OnTE,Chaudhry2019UsingHT,Buzzega2020DarkEF,Dam2022ScalableAO,VinciusdeCarvalho2022ClassIncrementalLV,Masum2023AssessorGuidedLF,lopez2017gradient,Shin2017ContinualLW} and architecture-based approach \cite{Xu2021AdaptivePC,Yoon2017LifelongLW,Ashfahani2021UnsupervisedCL,Pratama2021UnsupervisedCL,Li2019LearnTG,Rakaraddi2022ReinforcedCL}. Although the memory-based approaches generally outperform the other two approaches, they impose storage and privacy concerns. Recently, there is a trend toward a rehearsal-free continual learning approach benefiting from strong generalization power of pre-trained model (PTM) \cite{Wang2021LearningTP,Wang2022DualPromptCP,Masum2025VisionAL,Momeni2024ContinualLU,Zhou2024ContinualLW}. That is, these approaches fix the backbone network to avoid the CF problem while adapting to new tasks via parameter-efficient fine-tuning (PEFT) approaches such as prompts \cite{Wang2021LearningTP,Wang2022DualPromptCP,Masum2025VisionAL}, LoRA \cite{He2025CLLoRACL,Liu2025LoRASF} and adapter \cite{Fukuda2024AdapterMW}. Such an approach is effective because it involves a small number of trainable parameters and is powerful because it enjoys generalized features of the foundation model. Nevertheless, most of these efforts are devoted to addressing the class-incremental learning (CIL) problems \cite{Zhou2023ClassIncrementalLA} where each task features disjoint classes and a model is expected to handle all classes of all tasks in the testing phase. 

Domain Incremental Learning (DIL) is a CL sub-problem where the goal is to handle a sequence of varying domains. It differs from the CIL problem because each task is drawn from different domains while sharing the same label space \cite{vandeVen2019ThreeSF}. That is, a model is trained to be robust against various changes, such as data style shifts, data quality degradation, environmental changes, etc., rather than to recognize new classes. In \cite{Liu2024CompositionalPF}, Compositional Prompt (C-Prompt) is proposed as a rehearsal-free solution to the DIL problem, where it puts forward the notion of a domain-specific prompt pool. \cite{Wang2025DualCPRD} proposes the concept of dual-level concept prototypes (DualCP), which consists of coarse-grained and fine-grained prototypes for each class. The idea of SOYO is presented in \cite{Wang2025BoostingDI}, where the key idea lies in the Gaussian Mixture Compressor and Domain Feature Resampler. \cite{Zhou2024DualCF} offers the idea of dual consolidations in the feature level and in the classifier level. Although DIL has rapidly grown, existing works suffer from at least three bottlenecks: 1) they overlook shared knowledge across tasks. Notwithstanding that \cite{Liu2024CompositionalPF} also makes use of the global prompt to capture shared knowledge. We offer an alternative approach here, where the idea of low-rank adaptation (LoRA) is implemented. In addition, \cite{Liu2024CompositionalPF} imposes considerable complexities because of the use of domain-specific prompt pools; 2) they suffer from low parameter selection accuracies, leading to incorrect parameters being utilized for inferences. That is, existing approaches often apply the matching degree between the prompt key and query to select task-specific prompts during inferences, resulting in inaccurate parameter selections; 3) they adopt the linear or prototype-based classifiers having subpar generalization power.    

To correct this shortcoming, we propose continual knowledge consolidation low-rank adaptation (CONEC-LoRA) to deal with the DIL problems. CONEC-LoRA features task-specific LoRAs and task-shared LoRAs in which the first $l$ blocks are assigned to the task-shared LoRA, leaving the remainder $L-l$ blocks for the task-specific LoRA. Our parameter selection strategy for inference is driven by an auxiliary network predicting the domain ID of a testing sample. The auxiliary network is trained during the training process with the projection-based Gaussian mixture model (PGMM) to address the class imbalance problem caused by the absence of previous samples. The idea of the projection or transformation module is introduced to prevent the synthetic sample-bias problem. Another innovation lies in the different-depth network structure, where every layer is assigned a local classifier. The final prediction is drawn from a local classifier maximizing the logit. Last but not least, the concept of a stochastic classifier is integrated to improve the model's generalizations. Unlike the popular prototype-based classifier, the stochastic classifier benefits from the mean and variance vectors sampled from a distribution. This technique guarantees the presence of an infinite number of classifiers, promoting correct classifications.   

The key differences between our approach and \cite{He2025CLLoRACL} lie in the task-specific LoRA selection strategy, where the auxiliary network is implemented to select the task-specific LoRA rather than the similarity-based weighting scheme. Such a strategy guarantees the isolation of task-specific LoRAs, allowing for the retention of task-specific details, i.e., robust against the CF problem. By extension, \cite{He2025CLLoRACL} applies the prototype-based classifier, whereas ours is underpinned by the stochastic classifier. We also address the DIL problem here rather than the CIL problem. Note that the adapter fusion strategy in \cite{He2025CLLoRACL} undermines the domain-specific knowledge because all task-specific LoRAs are aggregated. Consequently, this issue leads to the use of orthogonality loss in their approach. In addition, such an approach incurs considerable complexities because multiple forward passes need to be carried out for all task-specific LoRA. On the contrary, our approach is much simpler than that because it predicts the correct task-specific LoRA followed by the network inference. This paper conveys at least four major contributions.
\begin{itemize}
    \item This paper proposes continual knowledge consolidation low-rank adaptation (CONEC-LoRA) for domain incremental learning (DIL) problems, where it features general LoRAs to extract cross-task knowledge and task-specific LoRAs to capture domain-specific knowledge. The building block of CONEC-LORA is constructed under a stacked pretrained transformer block of $L$ layers, where the first $l$ blocks are designated for the task-shared LoRA, while the remaining $L-l$ blocks are reserved for the task-specific LoRA. 
    \item A parameter selection strategy is proposed based on an auxiliary network predicting the domain label for the task-specific LoRAs for inferences. The auxiliary network receives an input of the frozen backbone and is trained with the projection-based GMM under a joint loss to address the class imbalance issue as a result of the absence of previous samples. The concept of the projection or transformation module is applied to random inputs of the GMM to prevent the synthetic sample bias problem. In addition, we introduce the ball-generator loss to prevent the synthetic sample bias problem. The auxiliary network is structured under a different-depth configuration, where each layer is connected to a local classifier with an early exit strategy to reduce computational complexities. The local classifier with the most confident prediction is selected for inferences.  
    \item The concept of the stochastic classifier is put forward for domain incremental learning (DIL) problems. This idea extends \cite{Paeedeh2024FewShotCI} devised for the CIL problem. This classifier is represented by learnable mean and covariance vectors, with the classifier weight defined as a distribution. This trick enables the creation of an arbitrary number of classifiers, increasing the chance of correct predictions.  
    \item The advantage of CONEC-LORA has been rigorously evaluated under four benchmark DIL problems. It is compared with prominent DIL algorithms where CONEC-LORA outperforms them by over $5\%$ margin. In addition, the source code of CONEC-LORA is made publicly available in the GitHub repository of the paper~\footnote{https://github.com/Naeem-Paeedeh/CONEC-LoRA}  for reproducibility and convenient further study.  
\end{itemize}  % the GitHub repository of the paper \href{https://github.com/Naeem-Paeedeh/CONEC-LoRA}{https://github.com/Naeem-Paeedeh/CONEC-LoRA}
The remainder of this paper is structured as follows: Section 2 discusses the related works; Section 3 outlines preliminaries, including a problem definition and basic concepts; Section 4 describes our algorithm, namely CONEC-LoRA; Section 5 offers our numerical study; and some concluding remarks are drawn in the last section of this paper. 

\section{Related Works}
Continual learning (CL) aims to address non-stationary learning problems, where a model cannot be fixed once deployed because its predictions quickly become outdated due to changing learning environments. That is, a CL agent is faced with never-ending environments under dynamic conditions \cite{chen2018lifelong,Parisi2018ContinualLL,Masana2020ClassIncrementalLS,Zhou2023ClassIncrementalLA}, resulting in the stability-plasticity dilemma. The CL problem can be divided into three sub-problems \cite{vandeVen2019ThreeSF}: task-incremental learning (TIL), class-incremental learning (CIL), and domain-incremental learning (DIL). The TIL and the CIL are inherently identical, where the difference lies in only the presence of task identifiers during the evaluation phase in the 
TIL. Therefore, the CIL is deemed more challenging than the TIL.  

\noindent\textbf{Class-Incremental Learning (CIL) problem} is formulated as a learning problem of sequentially arriving classes. That is, each task presents a set of classes disjoint across tasks. Learning a new task, thus, induces the CF problem because previously valid parameters are catastrophically erased with new ones, i.e., parameter drifts. The regularization-based approach \cite{Kirkpatrick2016OvercomingCF,Mao2021ContinualLV,Paik2019OvercomingCF,Zenke2017ContinualLT,aljundi2018memory,Li2019LearnTG,Schwarz2018ProgressC,Cha2020CPRCR} deals with the CF problem via integration of a regularization term to avoid the parameter drift problem. However, the regularization-based approach usually doesn't scale well for large-scale problems because it is difficult to find an overlapping region of all tasks. The architecture-based approach \cite{Xu2021AdaptivePC,Yoon2017LifelongLW,Ashfahani2021UnsupervisedCL,Pratama2021UnsupervisedCL,Li2019LearnTG,Rakaraddi2022ReinforcedCL} adds new components while isolating old components when learning new tasks to combat the CF problem. Nevertheless, the architecture-based approach usually calls for the existence of the task IDs, thus being incompatible with the CIL problem, otherwise, it requires old samples to be stored. The memory-based approach \cite{Chaudhry2018EfficientLL,Rebuffi2016iCaRLIC,Chaudhry2019OnTE,Chaudhry2019UsingHT,Buzzega2020DarkEF,Dam2022ScalableAO,VinciusdeCarvalho2022ClassIncrementalLV,Masum2023AssessorGuidedLF,lopez2017gradient,Shin2017ContinualLW} utilizes a memory buffer storing old samples. Old samples can then be interleaved with new samples for experience replay steps to prevent the CF problem. Although it is evident that the memory-based approach outperforms the other two approaches, it raises storage and privacy concerns. This issue motivates the development of rehearsal-free approaches benefiting from the strong generalization power of pretrained model (PTM) \cite{Zhou2024ContinualLW}. This approach is combined with the parameter-efficient fine-tuning (PEFT) approaches such as prompts \cite{Wang2021LearningTP,Wang2022DualPromptCP,Masum2025VisionAL}, LoRA \cite{He2025CLLoRACL,Liu2025LoRASF}, and adapter \cite{Fukuda2024AdapterMW}. Such an approach offers strong performances because the CF problem can be effectively alleviated to a very low level by freezing the backbone networks while only adjusting a small number of external parameters. Although there exist numerous contributions in the realm of the CIL problem, the DIL problem remains a relatively uncharted territory. 

\noindent\textbf{Domain Incremental Learning (DIL) problem} is unlike the CIL problem, where each task carries a unique domain while sharing the same label space. That is, a model is trained to be robust against various changes rather than to recognize new classes. \cite{Liu2024CompositionalPF} proposes the idea of compositional prompts, where each domain is assigned a prompt pool. In addition, the global prompt is injected and shared across all tasks. \cite{Wang2025DualCPRD} proposes the idea of dual prototypes, where coarse-grained and fine-grained prototypes are put forward. In \cite{Wang2025BoostingDI}, the concepts of Gaussian Mixture Compressor and Domain Feature Resampler are introduced to address inaccurate parameter selections in the DIL context. \cite{Zhou2024DualCF} proposes dual knowledge consolidations in the feature level as well as in the classifier level. The class imbalance problem in the DIL context is analyzed and addressed in  \cite{li2025addressing} by using the multi-expert concept. In \cite{xu2025componential}, the KA-prompt is proposed, where the key idea lies in the initialization of the new prompt with the most compatible previous prompt, thus promoting cross-domain knowledge. Nonetheless, we note at least three shortcomings in existing works: 1) they focus on the domain-specific knowledge while overlooking the domain-shared knowledge, resulting in sub-optimal performances. Note that \cite{Liu2024CompositionalPF} is based on the idea of the prompt, whereas our approach is constructed under the LoRA concept. In addition, \cite{Liu2024CompositionalPF} imposes prohibitive complexities due to the use of prompt pools for every domain; 2) existing approaches suffer from inaccurate parameter selection during inferences, significantly affecting the model's performance because wrong components are associated when producing an output. Although this problem is partially addressed in \cite{xu2025componential}, this approach relies on the greedy search technique, incurring prohibitive complexities; 3) existing approaches are built upon a linear or prototype-based classifier having limited generalization power. 

\section{Preliminaries}
\subsection{Problem Definition}

Domain-incremental learning (DIL) is formulated as a learning problem of sequentially arriving domains $\{\mathcal{D}_{1},\mathcal{D}_{2},...,\mathcal{D}_{B}\}$, where $B$ is the number of domains. Each domain comprises the training set $\mathcal{T}_{\text{tr}}^{b}$ and the testing set $\mathcal{T}_{\text{te}}^{b}$, where $\mathcal{D}_{b}=\{\mathcal{T}_{\text{tr}}^{b},\mathcal{T}_{\text{te}}^{b}\}$. When encountering the $b-th$ domain, a model is only presented with the training set of the $b-th$ domain $\mathcal{T}_{\text{tr}}^{b}$ with the absence of any training samples of the previous domains $\mathcal{T}_{\text{tr}}^{1\sim(b-1)}=\cup_{t=1}^{b-1}\mathcal{T}_{\text{tr}}^{t}$ and any exemplars of the previous domains leading to the catastrophic forgetting (CF) problem. The evaluation is performed against the testing samples of all already seen domains $\mathcal{T}_{\text{te}}^{1\sim b}=\cup_{t=1}^{b}\mathcal{T}_{\text{te}}^{t}$. The training set of the $b-th$ domain is formed by $N_b$ tuples $\mathcal{T}_{\text{tr}}^{b}=\{(x_i,y_i)\}_{i=1}^{N_b}$, where $x_i$ is an $i-th$ image and $y_i$ is its corresponding label. Suppose that $\mathcal{Y}_{b}$ presents the label set of the $b-th$ domain, each domain shares the same label set $\forall i\in [1,N_b],y_{i,b}\in\mathcal{Y}_{b}$, where $\forall b,b^{'}\in[1,B], \mathcal{Y}_{b}=\mathcal{Y}_{b^{'}}$ but features different characteristics such as styles and/or environmental changes. That is, there exists the presence of concept drifts $P(\mathcal{X},\mathcal{Y})_{b}\neq P(\mathcal{X},\mathcal{Y})_{b^{'}}$. No domain identifiers (IDs) are offered for inferences. In other words, a model is supposed to infer the domain IDs $b$ by itself. 

Our model $f_{\theta}=h_{\phi}\circ g_{\psi}$ relies on a pre-trained vision transformer (ViT) model, where $h_{\phi}:\mathcal{Z}\rightarrow\mathcal{Y}$ constitutes a classifier and $g_{\psi}:\mathcal{X}\rightarrow\mathcal{Z}$ denotes a feature extractor frozen during the training process. The training process of the $b-th$ domain is to adapt lightweight parameters $g_{W}$ inserted into the query and key projections of the attention module of all transformer blocks of the ViT backbone. Specifically, this is written as follows:
\begin{equation}
    z^{(\ell)}=g_{\psi}^{(\ell)}(x)+g^{(\ell)}_{W}(x),
\end{equation}
where $z^{(\ell)}$ is an embedded feature of layer $\ell$ for query or values. In addition, the classifier $h_{\phi}$ is defined as the stochastic classifier here. 

\subsection{Low-Rank Adaptation (LoRA)}
LoRA is a parameter-efficient fine-tuning (PEFT) approach \cite{Hu2021LoRALA,He2025CLLoRACL}, which is capable of adapting a foundation model to downstream tasks. The key idea is seen in the use of a pair of rank decomposition matrices, namely a down-projection matrix $B\in\Re^{r\times k}$ and an up-projection matrix $A\in\Re^{d\times r}$ with rank $r \ll \operatorname{min}(d,k)$. Hence, it achieves learning efficiency by only learning $r\times(d+k)$, which can be expressed as follows: 
\begin{equation}
    z = g_\psi(x) + \Delta W x,
\end{equation}
where $\Delta W= AB$. The low-rank matrices can then be attached to the transformer block. For DIL, these matrices can be set to be domain-specific $\{A_{b},B_{b}\}$. When adapting to the $b-th$ domain, only $\Delta W_{b}$ is learnable, leaving other parameters fixed, thus combating the CF problem. However, such na\"ive approach loses shared knowledge across domains. Besides, it risks inaccurate parameter selection during inference, which can lead to a loss of performance because no oracle for domain IDs is provided during the testing process. 

\subsection{Stochastic Classifier}
\begin{figure*}[t!]
    \centering
    \includegraphics[width=\linewidth]{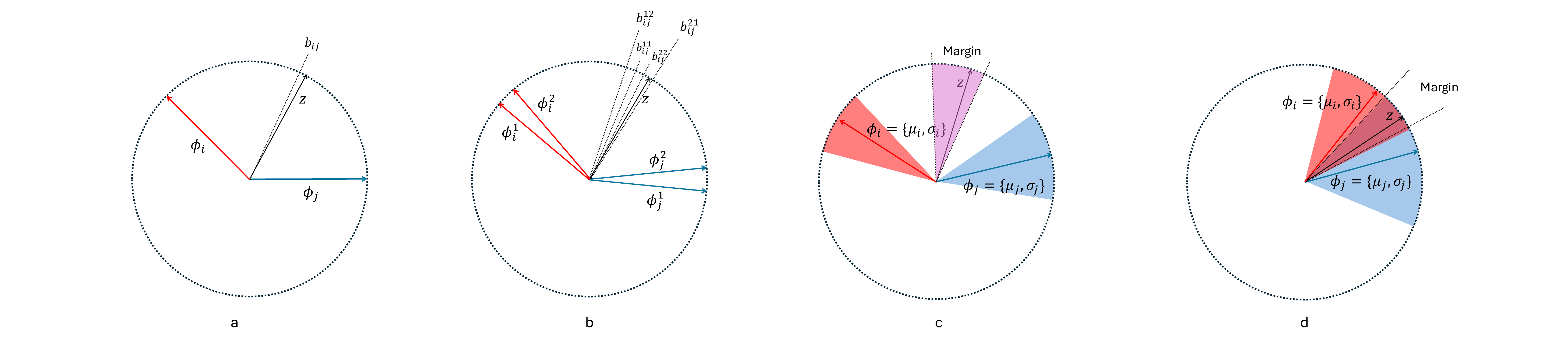}
    \caption{A stochastic classifier. a) The decision boundary for a single cosine classifier with two weight vectors for two classes. The feature vector belongs to the blue class. b) The illustration of the effect of having two weight vectors per class, resulting in four decision boundaries. c) It shows the effect of having an infinite number of weight vectors on the decision boundary. d) The stochastic nature of the classifier improves the feature separation of the model and classifier by resolving the confusion when the margin is unclear.}
    \label{fig:stochastic_classifier}
\end{figure*}

Given a classifier $h_{\phi}:\mathcal{Z}\rightarrow\mathcal{Y}$ mapping the latent space to the label space, the prototype-based classifier relies on a set of prototypes defined as an empirical mean of each class $\phi_{m}=\{\mu_{m}\}$. 
\begin{equation}
    \mu_{m}=\frac{1}{N_{m}}\sum_{i=1}^{N_{m}}\mathbb{I}_{y_{i}=m}z_{i},
\end{equation}
where $M$ denotes the dimension of the label space and $\mathbb{I}_{y_{i}=m}$ is an indicator function being true if $y_{i}=m$. Suppose that $\langle.\rangle$ stands for the cosine similarity, the predicted output of the prototype-based classifier is expressed as follows:
\begin{equation}
    P(y=m|x)=\frac{\exp({\eta \langle\phi_{m},z\rangle})}{\sum_{i=1}^{M}\exp({\eta\langle\phi_{m},z\rangle})},
\end{equation}
where $\eta$ is a temperature controlling the smoothness of a distribution. Such classification implies the use of a single static classifier, leading to suboptimal generalizations. The stochastic classifier \cite{Kalla2022S3CSS,Paeedeh2024FewShotCI} offers a different perspective where the classifier weights are sampled from a distribution $\phi_{m}=\{\mu_m,\sigma_{m}\}$, paving the way for the applications of multiple classifiers at once. The stochastic classifier is formalized:
\begin{equation}\label{stochastic}
\langle\phi_{m}, z\rangle \quad \text{s.t.} \quad \phi_{m}=\mu_{m}+\mathcal{N}(0,1)\odot\sigma_m,
\end{equation}
where $\mu_{m},\sigma_{m}$ are mean and variance vector automatically learned in an end-to-end manner and $\odot$ is an element-wise multiplication. Such a trick enables the creation of an arbitrary number of classifiers, thereby increasing the number of correct classifications. \cref{fig:stochastic_classifier} visualizes the stochastic classifier.

\section{Methodology}
\begin{figure*}[t!]
    \centering
    \includegraphics[width=0.9\linewidth]{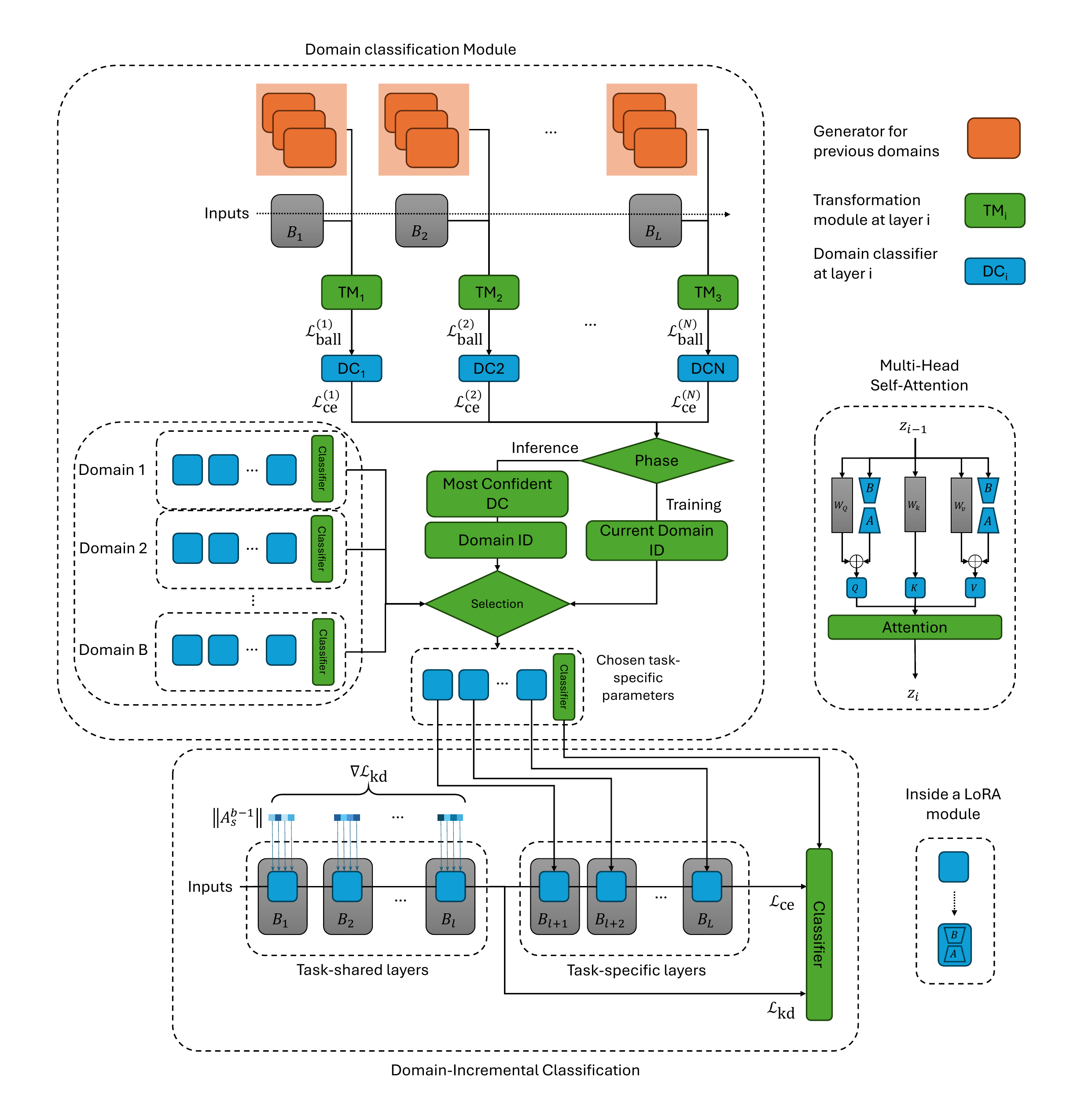}
    \caption{The architecture of the model and pipeline of CONEC-LoRA. CONEC-LoRA consists of two main stages. First, the input is given to the frozen backbone. The logits of the domain classifiers (DCs) are being calculated. The domain is detected based on the decision of the most confident domain classifier. Second, the domain-specific LoRAs are chosen for use in the domain-specific blocks. The gray rectangles represent the frozen layers of the ViT.}
    \label{fig:architecture}
\end{figure*}

This paper proposes continual knowledge consolidation low-rank adaptation (CONEC-LORA) built upon a synergy between task-shared LoRA and task-specific LoRA. That is, the first $l$ transformer blocks are attached with task-shared LoRAs while the remainder $L-l$ blocks are assigned with task-specific LoRAs. The classification decision is navigated with the stochastic classifier. For selecting the task-specific LoRAs during inference, the auxiliary network is applied to predict the domain label and is trained with the projection-based Gaussian mixture model (GMM) to address the class imbalance issue resulting from the absence of previous samples and any exemplars. The projection or transformation concept is implemented to cope with the synthetic sample bias problem while introducing a different-depth network structure for inference. An overview of CONEC-LORA's learning policy is outlined in \cref{fig:architecture}.

\subsection{Network Architecture}
Suppose the ViT backbone with $L$ blocks, the shared LoRAs $\{A_{s},B_{s}\}$ are inserted to the first $l$ blocks where $l\leq L$ while the remaining $L-l$ blocks are designated to the task-specific LoRA $\{A_{b},B_{b}\}$. Therefore, the output of the $i-th$ transformer block $z_{i}$ is written:
\begin{equation}
    z_{i}=g_{\psi_{i}}(z_{i-1})+\begin{cases}A_{s}^{i}B_{s}^{i}z_{i-1} \quad i\leq l\\ A_{b}^{i}B_{b}^{i}z_{i-1} \quad l<i\leq L\end{cases},
\end{equation}
where $z_{i-1}$ denotes the output of the previous transformer block $z_{0}=x$ while $g_{\psi_{i}}$ represents the $i-th$ transformer block frozen during the training process. $b$ is the domain label to be predicted during the inference. Note that early layers ($1$ to $l$) focus on general patterns to be shared across domains while deep layers ($l+1$ to $L$) capture domain-specific details. 

Motivated by \cite{He2025CLLoRACL}, the domain-shared LoRA applies a fixed random orthogonal down-projection matrix $B_{s}\in\Re^{r\times k}$ and a trainable up-projection matrix $A_{s}\in\Re^{d\times r}$ initialized as zeros. It satisfies the following property
\begin{equation}
    \Delta W x= A_{s}B_{s}x,\quad B_{s}B_{s}^{\top}=I,
\end{equation}
where $I\in\Re^{r\times r}$ is an identity matrix. \cite{Zhu2024AsymmetryIL} shows that adjusting $A$ is more effective than adjusting $B$. We also confirm that a random $B$ can compete to a fully trained one. First, a random matrix $M\in\Re^{r\times k}$ is generated from a normal distribution $\mathcal{N}(0,1)$. A random orthogonal matrix can be obtained via the singular value decomposition (SVD)~\cite{golub1965calculating}. 
\begin{equation}
    M=U\Sigma V^{\top}, B_{s}= UV_{r}^{\top},
\end{equation}
where $v_{r}$ is the first $r$ rows of $V$. In a nutshell, our effort is devoted to learn $A_{s}$ during the training process while keeping the random orthogonal matrix $B_{s}$ frozen.

CONEC-LoRA relies on the stochastic classifier during the training process, whose parameters are drawn from a distribution under the cosine classification strategy as per \eqref{stochastic}.
This strategy allows proper training process where $\phi_{m}=\{\mu_{m},\sigma_{m}\}$ are trained in an end-to-end fashion. Our investigation reveals that replacing each $\mu_m$ with the corresponding prototype after training leads to performance improvements at the inference phase. Note that the stochastic classifier with learnable $\mu_m$ must still be applied during the mini-batch training process, as using a prototype classifier during training results in performance degradation. In other words, the use of multiple classifiers as per the stochastic classifier is useful during the training process to boost the learning performance, but it may not surpass the flexibility of the prototype classifier during the inference phase.

Therefore, $\mu_m$ vectors of the stochastic classifier are replaced with prototypes for the inference phase.

\subsection{Loss Function}
CONEC-LORA is learned by minimizing a joint loss function consisting of the conventional cross-entropy loss and the knowledge distillation (KD) loss. 
\begin{equation}
\label{eq:DIL_optimization}
\mathcal{L}=\mathcal{L}_{\text{ce}}+\lambda_{1}\mathcal{L}_{\text{kd}},
\end{equation}
where $\lambda_{1}$ is a trade-off coefficient controlling the strength of the KD loss function. The KD strategy is omnipresent in the CL domain to combat the CF problem but this strategy is often too strict and leads to degraded plasticity. Hence, an early exit mechanism is implemented here where the KD approach is applied at the transition point between task-shared and task-specific LoRA, i.e., the $l-th$ block. Specifically, using a local classifier $h_{\phi}$, the KD mechanism is done by extracting the $[\text{CLS}]$ token representation through the shared LoRAs, i.e., $1$ to $l$, for both current domain $z_{b}^{l}[\text{CLS}]$ and previous domain $z_{b-1}^{l}[\text{CLS}]$. 
\begin{equation}
    \label{eq:loss_kd}
    \mathcal{L}_{\text{kd}}=\sum_{i\in D_{b}}s_{b-1,i}^{\tau}\log{s_{b,i}^{\tau}},
\end{equation}
where $s_{b}^{\tau}=\operatorname{Softmax}(h_{\phi}^{b}(z_{b}^{l}[\text{CLS}])/\tau)$ and $\tau=2$ is a temperature. Because $A_{s}$ is initialized as zero and learned during the training process, its norms mirror the contribution of each element, implying which parts need further tuning. This fact enables redistribution of the gradient from $\mathcal{L}_{\text{kd}}$ based on the weight importance in the previous training session. Suppose that $||a_{s,j}^{b-1}||_{2}$ denotes the $L_2$ norm of the $j-th$ element of the up-projection matrix $A_{s}^{b-1}$ of the previous domain, the gradient of $A_{s}^{b}$ of the current domain is redistributed:
\begin{equation}
    \label{eq:grad_red}
    \nabla_{A_{s}^{b}}\mathcal{L}_{\text{kd}}=\nabla_{A_{s}^{b}}\mathcal{L}_{\text{kd}}\odot\sigma{(\{||a_{s,j}^{b-1}||\}_{j=1}^{d})},
\end{equation}
where $\odot$ is Hadamard product, $\sigma(w)=d\times w/\sum_{i=1}^{d}w_{i}$ is the dimension-preserving normalization function, and $d$ is the embedding dimension. This redistribution strategy assures that each element is adaptively adjusted, i.e., the essential dimension for the previous task is preserved while allowing other dimensions to move freely in respect to the new task.

\subsection{Auxiliary Domain Classifiers}

CONEC-LORA is guided by an auxiliary network that predicts the domain label during inference. The predicted domain label is thus exploited to select the task-specific LoRAs without the CF problem. Since the auxiliary network is trained, it is expected to be more reliable than a non-parametric approach, such as the proposed method in \cite{Paeedeh2024FewShotCI} that uses the Mahalanobis distance. The representations of the intermediate layers of \Glspl{nn} are more robust to the changes during the continual learning and suffer less from forgetting \cite{Szatkowski2024ImprovingCL}. Furthermore, the classification can potentially be finished on earlier layers with even higher accuracy. In CONEC-LoRA, we keep the backbone frozen without utilizing the LoRAs and train independent domain classifiers at different levels of abstraction in the selected layers of a ViT. This ensures that the domain classification module would not be affected during the weight updates.\par

The domain label is predicted by a local linear classifier for every layer. That is, every layer is assigned with a local classifier $h_{\phi_{\text{aux}}^{l}}(.)$. We also insert a transformation module constructed as a two-layer MLP  $\kappa_{\gamma}^{l}(.)$ to prevent the synthetic sample bias problem, where $\gamma^{l}$ denotes the parameters of the transformation module of the $l-th$ layer. Given the embedding of the $l-th$ layer with the frozen backbone network and without any LoRAs $z_{l}$, the local output of the $l-th$ layer of the domain classification module is expressed as follows: 
\begin{equation}
    b=h_{\phi_{\text{aux}}^{l}}(\kappa_{\gamma^{l}}(z_{l})), l\in[1,L],
\end{equation}
where $b\in[0,1]$ denotes the output logit of the $l-th$ local classifier and $L$ is the number of layers. The aggregation is performed by applying the maximum operator to each local classifier in every layer. 
\begin{equation}
    \label{eq:confidence}
    \hat{b}=\arg\max_{1\leq l\leq L}h_{\phi_{\text{aux}}^{l}}(\kappa_{\gamma^{l}}(z_{\text{l}})),
\end{equation}
where $\hat{b}\in B$ and $B$ is the number of domains. In other words, the local classifier having the most confident prediction is selected for inference. The different-depth network structure is designed to benefit from the intermediate representation, making it more robust to dynamic conditions than a single classifier \cite{Szatkowski2024ImprovingCL}. During the inference, we also implement the early exit mechanism to relieve the computational burden. That is, we set a confidence threshold $\varsigma$ and examine whether the confidence of local classifiers exceeds the threshold. The earliest layer passing the threshold is selected for inference. This strategy avoids going through all layers for inference. 
\begin{equation}\label{confidence}
    h_{\phi_{\text{aux}}^{l}}(\kappa_{\gamma^{l}}(z_{\text{l}}))\geq\varsigma\rightarrow\hat{b}=\arg\max_{b}h_{\phi_{\text{aux}}^{l}}(z_{\text{l}})
\end{equation}
The earliest layer satisfying \eqref{confidence} is selected for inference, or the early exit mechanism is implemented. Nonetheless, the training process of the auxiliary network is non-trivial since there exists a class imbalance problem between the current domain and the previous domains. As with \cite{Wang2025BoostingDI}, we model the previous data distributions using a mixture of $C$ multivariate Gaussian distributions $\mathcal{N}(\nu, \cov)$.

\begin{equation}
    \mathcal{N}(z|\nu,\cov)=\frac{1}{(2\pi)^{\frac{d}{2}}|\cov|^{\frac{1}{2}}}\exp({-\frac{1}{2}(z-\nu)^{\top}\cov^{-1}(z-\nu)}),
\end{equation}
where $d$ is the dimension of embeddings. That is, the mean and covariance matrix of every domain $(\nu_{b}, \cov_{b})$ in the embedding space $\mathcal{Z}$ is calculated for each one of the $C$ components.

The probability density function of the mixture of the Gaussian distributions can be defined as:

\begin{equation}
    p(z) = \sum_{c=1}^{C} \omega_c \mathcal{N}(z|\nu_c,\cov_c),
\end{equation}
where $0\leq \omega_c \leq 1$ and $\sum_{c=1}^{C}\omega_c=1$. The GMM parameters are trained with the Expectation-Maximization (EM) algorithm \cite{bishop2006pattern}. By storing the $\{\omega_c,\nu_c,\cov_c\}_{c=1}^{C}$, we can generate embeddings of the previously seen domain from the GMMs:

\begin{align}
    q_l \sim \operatorname{Categorical}(\omega_l),\\
    \hat{z_l}|q_l\sim \mathcal{N}(\nu_{q_l}^l,\cov_{q_l}^l).
\end{align}

Nevertheless, this na\"ive sampling strategy causes the synthetic sample bias problem. To remedy this problem, the transformation module $\kappa_{\gamma^{l}}(.)$ is integrated. Specifically, we feed $\{\check{z}_\ell\} = \{\hat{z}_\ell\} \cup \{z_\ell\}$ set to the transformation module. For $\check{z}_{l}^i \in \{\check{z}_\ell\}$ we calculate $\hat{\hat{z}}_{l}^i=\kappa_{\gamma^{l}}(\check{z}_{l}^i)$, where $\hat{\hat{z}}_{l} \in\Re^{d}$. The ball-generator loss \cite{Masum2023FewShotCL} is introduced and inspired by the triplet and hinge losses.

\begin{equation}
\mathcal{L}_{\text{ball}}^{l}=\sum_{\hat{\hat{z}}_{l}\in \bar{y}_{i},\hat{\hat{z}}_{l} \notin \bar{y}_{j}}\max\{0,d(\hat{\hat{z}}_{l},\mu_{i})+r-d(\hat{\hat{z}}_{l},\mu_{j})\},
\end{equation}
where $r$ is a predefined margin constant, $\bar{y}$ is the domain label, and $d(.)$ is a distance measure. The goal of the ball-generator loss is to pull the synthetic sample towards and close to its center while keeping it away from other centers. Hence, the local auxiliary network $h_{\phi_{\text{aux}}^{l}}(.)$ of the $l-th$ layer is trained to minimize the following joint loss. 
\begin{algorithm}[!t]
    \caption{CONEC-LoRA: Training the LoRAs and classifiers}
    \label{alg:conec_lora_1}
    \begin{algorithmic}[1]
        \REQUIRE Training sets $\{\mathcal{T}_{\text{tr}}^{b}\}_{b=1}^B$ for $B$ domains, where $\mathcal{T}_{\text{tr}}^{b}=\{(x_i,y_i)\}_{i=1}^{N_b}$, backbone $g_{\psi}$ with $L$ layers, classifiers set $\{h_{\phi_b}\}_{b=1}^{B}$, task-shared LoRAs $\{g_{W_\ell^{s}}\}_{\ell=1}^l$, task-specific LoRAs $\{g_{W_\ell^{b}}\}_{\ell=l+1}^{L}$, $M$ as the number of epochs, $B$ domains, and $\mathcal{C}$ classes for every domain.
        \STATE {$\blacktriangleright$ \textbf{Training}}:
        \FOR{each $b \in \{1, 2, \dots, B\}$}
            % \STATE \COMMENT{Training the LoRAs}

            \STATE $h_{\text{temp}} \gets \textsc{Temporary Stochastic Classifier}$
            % \STATE $g_W^s$ are the task-shared LoRAs and $g_W^b$ the task-specific LoRAs
            \FOR{each $i \in \{1, 2, \dots, M\}$}
                % \STATE $\mathcal{L}_{\text{kd}}^j = 0$
                \FOR{each $(x_j,y_j) \in \mathcal{T}_{\text{tr}}^{b}$}
                    % \STATE $z_0 \gets x_j$
                    \STATE $z_j \gets g_\psi(x_j; \{g_{W_\ell^s}\}_{\ell=1}^l, \{g_{W_\ell^b}\}_{\ell=l+1}^{L})$
                    \STATE $z_{j} \gets h_{\text{temp}}(z_j)$
                    \STATE $\mathcal{L}_{\text{ce}} \gets \operatorname{Loss_{CE}}(z, y_j)$   \COMMENT{Classification loss}
                    
                    % \COMMENT{The KD-Loss}
                    \STATE $z_j^b \gets x_j$
                    \IF{$b > 1$}
                        \STATE $z_j^b \gets g_\psi(x_j; \{g_{W_\ell^s}^b\}_{\ell=1}^l)$
                        \STATE $z_j^{b-1} \gets g_\psi(x_j; \{g_{W_\ell^s}^{b-1}\}_{\ell=1}^l)$
                        
                        \STATE $\mathcal{L}_{\text{kd}}^j \gets \mathcal{L}_{\text{kd}}^j + s_{b-1,j}^{\tau}\log{s_{b,j}^{\tau}}$ \COMMENT{\cref{eq:loss_kd}}
                        \FOR{each $\ell \gets \{1, 2, \dots, l\}$}
                            % \STATE \COMMENT{\cref{eq:grad_red}}
                            \STATE $\nabla_{A_{s}^{b,\ell}}\mathcal{L}_{\text{kd}}^j=\nabla_{A_{s}^{b,\ell}}\mathcal{L}_{\text{kd}}^j\odot\sigma{(\{||a_{s,k}^{b-1,\ell}||\}_{k=1}^{d})}$
                        \ENDFOR
    
                        % \STATE $z_{b-1} \gets g_\psi(x_j; g_{W_{b-1}}^s)$
                    \ENDIF
                \ENDFOR
                \STATE Minimize the $\mathcal{L} = \mathcal{L}_{\text{ce}}(z, y) + \lambda_1 \mathcal{L}_{\text{kd}}$ \COMMENT{\cref{eq:DIL_optimization}}
            \ENDFOR
            \STATE \COMMENT{Setting the weight of the classifier to prototypes}
            % \ENSURE Prototypes $\mu = \{\mu_c\}_{c \in \mathcal{C}}$
            \FOR{each $c \in \mathcal{C}$}
                \STATE $S = \{x_j | y_j = c\}, (x_j, y_j) \in \mathcal{T}_{\text{tr}}^{b}$
                \STATE $\mu_c \gets \frac{1}{|S_c|} \sum_{x \in S_c}g_\psi(x_j; \{g_{W_\ell^s}\}_{\ell=1}^l, \{g_{W_\ell^b}\}_{\ell=l+1}^{L})$ \COMMENT{Prototypes}
                \STATE $\sigma_c \gets h_{\text{temp}}$
                \STATE $h_{\phi_b} \gets (\mu_c, \sigma_c)$
            \ENDFOR
        
        \ENDFOR
    \end{algorithmic}
\end{algorithm}

\begin{algorithm}[!t]
    \caption{CONEC-LoRA: Training the Domain Classification Module}
    \label{alg:conec_lora_2}
    \begin{algorithmic}[1]
        \REQUIRE Training sets $\{\mathcal{T}_{\text{tr}}^{b}\}_{b=1}^B$ for $B$ domains, where $\mathcal{T}_{\text{tr}}^{b}=\{(x_i,y_i)\}_{i=1}^{N_b}$, backbone $g_{\psi}$ with $L$ layers, domain classifiers set $\{h_{\phi_{\text{aux}}^{\ell}}\}_{\ell=1}^{L}$, $M$ as the number of epochs, $\mathcal{C}$ classes for every domain, a set of transformation modules $\{\kappa_{\gamma}^{\ell}\}$, $C$ components for GMMs, and $N$ as the number of synthetic embeddings to generate.
        \STATE {$\blacktriangleright$ \textbf{Training the Domain Classification Module}}:
        \FOR{each $b \in \{1, 2, \dots, B\}$}
            \STATE \COMMENT{Obtaining the embeddings for every layer of the current domain}
            
            % Calculating the embeddings, obtaining the centers for the ball-generator loss
            \FOR{each $x_j \in \mathcal{T}_{\text{tr}}^{b}$}
                \STATE $z_0^j \gets x_j$
                \FOR{each $\ell \in \{1, 2, \dots, L\}$}
                    % \STATE \COMMENT{Embedding vectors from the frozen backbone for every layer without LoRAs}
                    \STATE $z_\ell^j \gets g_{\psi_{\ell-1}}(z_{\ell-1}^j)$  \COMMENT{Without LoRAs}
                    \STATE $\bar{y}_{b,j} \gets b$
                \ENDFOR
            \ENDFOR
            
            \STATE \COMMENT{Calculating the centers for ball-generator loss}
            \FOR{each $\ell \in \{1, 2, \dots, L\}$}
                \STATE $\mu_\ell^{b} \gets \frac{1}{|Z_\ell|} \sum_{z \in \{z_\ell\}}z$ \COMMENT{Center for layer $\ell$}
            \ENDFOR

            \STATE \COMMENT{Finding the GMM parameters for the following tasks}
            \FOR{each $\ell \in \{1, 2, \dots, L\}$}
                \STATE $\{ \omega_c^{b,\ell}, \nu_c^{b,\ell}, \cov_c^{b,\ell} \}_{c=1}^{C} \gets \operatorname{GMM_{EM}}(Z_{\ell})$
            \ENDFOR
            
            \STATE \COMMENT{Generating the synthetic embeddings}
            
            \FOR{each $d \in \{1, 2, \dots, b - 1$\} }
                \FOR{each $k \in \{1, 2, .., N\}$}
                    \STATE $q_{d,k} \sim \operatorname{Categorical}(\omega^{d})$
                    \STATE $\hat{z}_{d,k}|q_{d,k} \sim \mathcal{N}(\nu_{q_{d,k}}^{d},\cov_{q_{d,k}}^{d})$
                    \STATE $\bar{y}_{d,k} \gets d$
                \ENDFOR
            \ENDFOR

            \FOR{each $m \in \{1, 2, \dots, M\}$}
                \FOR{each $\ell \in \{1,2, \dots, L\} $}
                    \STATE \COMMENT{Calculation of the ball-generator loss}
                        \IF{$b > 1$}
                            \STATE $\{ \check{z}_\ell \} \gets \{ \hat{z}_\ell \} \cup \{ z_\ell \}$
                            \FOR{each $\check{z}_{\ell}^j \in \{\check{z}_{\ell} \}$}
                                \STATE $\hat{\hat{z}}_{\ell}^j \gets \kappa_{\gamma^{\ell}}( \check{z}_{\ell}^j  )$
                            \ENDFOR
                            \STATE $\mathcal{L}_{\text{ball}}^{\ell}=\sum_{\hat{\hat{z}}_{\ell} \in \bar{y}_{i},\hat{\hat{z}}_{\ell}\notin \bar{y}_{j}}\max\{0,d(\hat{\hat{z}}_{\ell},\mu_\ell^i)+r-d(\hat{\hat{z}}_{\ell},\mu_\ell^j)\}$
                            \STATE $\mathcal{L}_{\text{aux}}^{\ell} \gets \mathcal{L}_{\text{ce}}^{\ell}(h_{\phi_{\text{aux}}^{\ell}}(\hat{\hat{z}}_\ell)) +\lambda_2\mathcal{L}_{\text{ball}}^{\ell}$
                        \ELSE
                            \STATE $\mathcal{L}_{\text{aux}}^{\ell} \gets \mathcal{L}_{\text{ce}}^{\ell}(h_{\phi_{\text{aux}}^{\ell}}(z_\ell))$
                        \ENDIF
                \ENDFOR
                \STATE Minimize the $\mathcal{L}_{\text{aux}}$
            \ENDFOR
            
        \ENDFOR
    \end{algorithmic}
\end{algorithm}

\begin{equation}
\mathcal{L}_{\text{aux}}^{l}=\mathcal{L}_{\text{ce}}^{l}+\lambda_2\mathcal{L}_{\text{ball}}^{l},
\end{equation}
where $\lambda_{2}$ is a tradeoff coefficient controlling the strength of the ball generator loss. Note that $\mathcal{L}_{\text{ce}}^{l}$ is subject to both the current samples $z\sim\mathcal{D}_{b}$ and the synthetic samples $\hat{z_l}|q_l\sim \mathcal{N}(\nu_{q_{d,k}}^l,\cov_{q_{d,k}}^l), q_l \sim \operatorname{Categorical}(\omega_l)$ representing the previous domains. The pseudo-codes of different phases of the CONEC-LoRA are shown in \cref{alg:conec_lora_1,alg:conec_lora_2,alg:conec_lora_3}.

\begin{algorithm}[!t]
    \caption{CONEC-LoRA: Inference phase}
    \label{alg:conec_lora_3}
    \begin{algorithmic}[1]
        \REQUIRE Test set $\mathcal{T}_{\text{te}}^{1\sim b}=\cup_{t=1}^{b}\mathcal{T}_{\text{te}}^{t}$, backbone $g_{\psi}$, classifiers set $\{h_{\phi_b}\}_{b=1}^{B}$, domain classifiers set $\{h_{\phi_{\text{aux}}^{\ell}}\}_{\ell=1}^{L}$, a set of transformation modules $\{\kappa_{\gamma}^{\ell}\}$, task-shared LoRAs $\{g_{W_\ell^{s}}\}_{\ell=1}^l$, task-specific LoRAs $\{g_{W_\ell^{b}}\}_{\ell=l+1}^{L}$, $\mathcal{C}$ classes for every domain, and threshold value $\tau$.
        
        \STATE {$\blacktriangleright$ \textbf{Inference phase}}:
        \FOR{each $x_i \in \mathcal{T}_{\text{te}}^{b}$}
            \STATE \COMMENT{Obtaining the embeddings for every layer}
            
            \STATE $z_0^i \gets x_i$
            \FOR{each $\ell \in \{1, 2, \dots, L\}$}
                \STATE \COMMENT{Embedding vectors from the frozen backbone for every layer without LoRAs}
                \STATE $z_\ell^i \gets g_{\psi_{\ell-1}}(z_{\ell-1}^i)$
                \STATE $\tilde{z}_\ell^i \gets h_{\phi_{\text{aux}}^{\ell}}(\kappa_{\gamma^{\ell}}( z_\ell^i  ))$
                \STATE \COMMENT{Confidence of every domain classifier}
                \STATE $\rho_\ell^i \gets \max\tilde{z}_\ell^i$
            \ENDFOR

            \STATE $S_i \gets \{\ell | \rho_\ell^i \ge \tau\}, 1 \le \ell \le L$
            
            \IF{$S \neq \emptyset$}
                \STATE \COMMENT{The index of the first classifier that meets the criterion}
                \STATE $\gamma_i \gets \min (S_i)$
            \ELSE
                \STATE \COMMENT{The index of the most confident classifier}
                \STATE $\gamma_i \gets \underset{\ell}{\arg\max} (\rho_\ell^i)$
            \ENDIF
            
            \STATE $\hat{b}_i \gets \arg\max (\rho_{\gamma^i}^i)$ \COMMENT{Predicted domain label}

            \STATE $z_i \gets g_\psi(x_i; \{g_{W_\ell^s}\}_{\ell=1}^l, \{g_{W_\ell^{\hat{b}_i}}\}_{\ell=l+1}^{L})$
            \STATE $z_{i} \gets h_{\phi_{\hat{b}_i}}(z_i)$
            \STATE $\hat{y}_i \gets \arg\max (z_{i})$
        \ENDFOR
    \end{algorithmic}
\end{algorithm}

\subsection{Time Complexity Analysis}

The proposed method requires two forward passes during training or inference. In one pass, we forward the inputs without LoRA modules to obtain the embeddings for the intermediate layers for the intermediate domain classifiers. In another pass, we should obtain the network's outputs by using both the shared and domain-specific LoRA modules. The crucial parts of the network that their calculations primarily affect in terms of time complexity include the two forward passes of the ViT and LoRA modules, the auxiliary components of the network, and fitting the GMM. In the following, we first calculate the time complexity of those components. Next, we calculate the overall complexity. \par

The major components of ViT calculations are the MLPs and attention mechanisms within each block. In a ViT with the embedding dimension of $d$ and sequence length of $s$, the query, key, and value projections require $O(s d^2)$. The attention mechanism requires $O(ds^2 + sd^2)$ operations. If a weight matrix in one layer of the MLP is $\tilde{h}_{\ell - 1} \times \tilde{h}_{\ell}$ and the width of an input is $n_0$, an MLP with $L_{\text{MLP}}$ layers requires $O\bigl(s \sum_1^{L_{\text{MLP}}} (\tilde{h}_{\ell - 1} \tilde{h}_{\ell})\bigr)$. Since a shallow two-layer MLP is commonly used in the ViTs, the time complexity becomes $O(sd\tilde{h}_1)$. Each LoRA adapter has the same time complexity as a shallow two-layer MLP. Two LoRA adapters for the Q and V with rank $r$ add $O(srd)$ operations ($r \ll d$) overhead to each block of the ViT.

Therefore, the total time complexity of a ViT with LoRAs is $O\left(Ls (ds + d^2 + d\tilde{h}_1 + rd)\right)$ or an $L$-layer ViT. \par

In GMM calculations, the most complex operations are the Mahalanobis distance calculations and matrix inversion. If the number of samples is $n$ and the number of components is $c$, the Mahalanobis distance operations are of $O(ncd^2)$. The matrix inversion requires $O(cd^3)$. However, since $c$ is usually small (2 in our experiments) and the dimension of embeddings in the network is fixed, this operation is not affected by the number of samples; therefore, it is negligible. Overall, the time complexity of the GMM is $O(I_\text{G}Lncd^2)$, where $I_{\text{G}}$ represents the maximum number of iterations for fitting the GMM. The GMM training occurs only during the training.\par

Finally, there are $L$ MLPs as transformation modules and domain classifiers after each layer of the ViT. Each MLP as a transformation module with $L_{\text{TM}}$ layers requires $O\bigl(s \sum_1^{L_{\text{TM}}} ( \hat{h}_{\ell - 1} \hat{h}_{\ell})\bigr)$ if a weight matrix in an MLP layer is $ \hat{h}_{\ell-1} \times \hat{h}_{\ell}$.
Moreover, each domain classifier requires $O(sd^2)$ operations. Therefore, the total time complexity of the domain classification phase (parameter selection) is of $O\Bigl(Ls \bigl(ds + d^2 + \sum_1^{L_{\text{TM}}} (\hat{h}_{\ell - 1} \hat{h}_{\ell})\bigr)\Bigr)$.\par

Considering all the complexities of the components, with a batch size of $B$, $I_\text{TC}$ iterations for training for the classification, and $I_\text{TD}$ iterations for the domain classification phase of the training, the time complexity of the training is 
\begin{equation}
\begin{split}
O\Bigl(BL \bigl( (I_\text{TC} + I_\text{TD}) (ds^2 + sd^2 + sd\tilde{h}_1) + \\
I_\text{TC}srd + I_\text{TD} s \sum_1^{L_{\text{TM}}} (\hat{h}_{\ell - 1} \hat{h}_{\ell})\bigr) + \\
I_\text{G}Lncd^2\Bigr),
\end{split}
\end{equation}
and the time complexity of the inference for a sample is 
\begin{equation}
O\Bigl( Ls \bigl(ds + d^2 + d\tilde{h}_1 + \sum_1^{L_{\text{TM}}} (\hat{h}_{\ell - 1} \hat{h}_{\ell})\bigr) \Bigr).
\end{equation}

\section{Experiments}

\subsection{Datasets}

We evaluate the effectiveness of CONEC-LoRA on four established DIL datasets: DomainNet~\cite{peng2019moment}, CORe50~\cite{lomonaco2017core50}, CDDB-Hard~\cite{li2023continual}, and Office-Home \cite{Venkateswara2017DeepHN}. These datasets are executed with a random seed across five domain orders. The final numerical results are taken from the average across five different domain orders. 

\subsection{Baseline Methods}

We compare our method comprehensively with DCE~\cite{li2025addressing}, DualCP \cite{Wang2025DualCPRD}, S-iPrompt~\cite{Wang2022SPromptsLW}, CODA-Prompt~\cite{Smith2022CODAPromptCD}, RanPAC~\cite{McDonnell2023RanPACRP}, EASE~\cite{Zhou2024ExpandableSE}, DualPrompt~\cite{Wang2022DualPromptCP}, L2P~\cite{Wang2021LearningTP}, and SimpleCIL~\cite{zhou2025revisiting}. Moreover, we report the results for the replay-based method, including MEMO~\cite{zhou2022model}, iCaRL~\cite{rebuffi2017icarl}, and Replay~\cite{ratcliff1990connectionist}. 

\subsection{Implementation Details}
The CDDB, Office-Home, and CORe50 experiments are conducted on an NVIDIA RTX 4090 GPU, while the DomainNet experiments are performed on an RTX A5000 GPU. The images are resized to 224x224 pixels for all experiments. The transformation modules are two-layer MLPs with a hidden layer dimension of 1024. Regarding to the hyper-parameters, the rank for LoRA matrices is set to $8$, $\lambda_1$ and $\lambda_2$ are set to 5 and 2, respectively. The margin is set to $1$. $\varsigma$ is set to $0.9$. The learning rate for the LoRAs and the temporary classifier is set to $0.02$. The learning rates for the domain classifiers and transformation modules are set to $\num{2e-3}$ and $\num{1e-4}$, respectively. We use the SGD optimizer for training the model with a batch size of $64$. The first 6 blocks of the network are dedicated to the task-shared LoRAs, and the next 6 layers are utilized for the task-specific LoRAs.

\subsection{Numerical Results}
\begin{table*}[t]
	\caption{\small Comparison of the average and last accuracies of different methods.
		The ViT-B/16 IN1K is used as a backbone. Other backbones are mentioned after the method name. Methods with $\dagger$ indicate implemented with exemplars (10 per class). Note that since the train and test domains of the CORe50 dataset do not overlap, the ground truth domain labels are not available, and the accuracy with the oracle cannot be calculated.
	}
    \label{tab:DIL_accuracies}
	% \vspace{-3mm}
	\centering
	\resizebox{0.95\textwidth}{!}{
		\begin{tabular}{ll  % @{}
                        *{4}{S S}
                        }
                \toprule
                  &
                  &
                  \multicolumn{2}{c}{Office-Home} &
                  \multicolumn{2}{c}{DomainNet} &
                  \multicolumn{2}{c}{CORe50} &
                  \multicolumn{2}{c}{CDDB-Hard} \\
                \cmidrule(lr){3-4}\cmidrule(lr){5-6}\cmidrule(lr){7-8}\cmidrule(lr){9-10}
                  Method & Venue & Avg & Last & Avg & Last & Avg & Last & Avg & Last \\
                \midrule
			Fine-tune~\cite{Zhou2024DualCF} & CVPR 2025 & 78.32{\tiny{$\pm$3.28}} & 76.16{\tiny{$\pm$1.39}} & 28.17{\tiny{$\pm$6.47}} & 38.82{\tiny{$\pm$7.65}} & 75.44{\tiny{$\pm$1.68}} & 76.19{\tiny{$\pm$2.36}} & 52.08{\tiny{$\pm$1.35}}  & 50.11{\tiny{$\pm$1.62}} \\
			
			Replay$^\dagger$~\cite{ratcliff1990connectionist} & Psychology Review & 84.23{\tiny{$\pm$2.31}} & 83.75{\tiny{$\pm$0.68}} & 64.78{\tiny{$\pm$2.98}} & 61.16{\tiny{$\pm$1.19}} & 85.56{\tiny{$\pm$0.38}} & 92.21{\tiny{$\pm$0.63}} & 66.91{\tiny{$\pm$18.0}}  & 63.21{\tiny{$\pm$11.6}}  \\
			
			iCaRL$^\dagger$~\cite{rebuffi2017icarl} & CVPR 2017 & \val{83.2}{3.2} & \val{83.5}{0.7} & \val{58.4}{5.3} & \val{55.9}{1.7} & \val{71.0}{3.7} & \val{76.0}{2.7} & \val{57.5}{9.9} & \val{54.4}{9.9} \\     % From DCE paper
			MEMO$^\dagger$~\cite{zhou2022model} & ICLR 2022 & \val{78.7}{3.7} & \val{79.1}{1.2} & \val{57.8}{6.7} & \val{56.4}{1.4} & \val{66.0}{2.7} & \val{68.2}{1.7} & \val{66.0}{2.7} & \val{68.2}{1.7} \\     % From DCE paper
			
			% \midrule
			
			SimpleCIL~\cite{zhou2025revisiting} & IJCV 2025 & \val{75.2}{4.9} & \val{76.2}{0.0} & \val{41.1}{6.8} & \val{40.6}{0.0} & \val{62.5}{1.6} & \val{67.2}{0.0} & \val{65.5}{3.2} & \val{64.1}{1.7} \\     % From DCE paper
			
			L2P~\cite{Wang2021LearningTP} & CVPR 2022 & \val{78.7}{4.2} & \val{80.5}{0.6} & \val{48.5}{6.8} & \val{45.2}{2.3} & \val{72.3}{1.3} & \val{81.7}{0.5} & \val{67.3}{5.3} & \val{65.0}{6.5} \\     % From DCE paper
			
			DualPrompt~\cite{Wang2022DualPromptCP} & ECCV 2022 & \val{77.2}{3.1} & \val{79.1}{0.6} & \val{52.3}{9.7} & \val{50.9}{4.3} & \val{71.0}{4.5} & \val{77.7}{1.0} & \val{66.9}{5.8} & \val{65.6}{2.6} \\     % From DCE paper
			
			CODA-Prompt~\cite{Smith2022CODAPromptCD} & CVPR 2023 & \val{82.4}{3.8} & \val{83.3}{0.3} & \val{47.6}{6.1} & \val{45.1}{1.2} & \val{72.8}{1.2} & \val{81.4}{1.1} & \val{67.9}{6.5}  & \val{66.6}{2.8} \\     % From DCE paper
			
			EASE~\cite{Zhou2024ExpandableSE} & CVPR 2024 & 81.16{\tiny{$\pm$3.52}} & 76.33{\tiny{$\pm$2.16}} & 50.50{\tiny{$\pm$2.27}} & 43.72{\tiny{$\pm$1.70}} & 86.30{\tiny{$\pm$0.04}}  & 87.02{\tiny{$\pm$1.21}} & 67.78{\tiny{$\pm$2.44}}  & 64.96{\tiny{$\pm$8.36}} \\
			
			RanPAC~\cite{McDonnell2023RanPACRP} & NeurIPS 2023 & \val{83.4}{3.8} & \val{83.3}{0.3} & \val{57.8}{5.5} & \val{56.1}{0.6} & \val{76.7}{1.3} & \val{78.4}{1.7} & \val{61.7}{4.3} & \val{62.5}{2.8} \\     % From DCE paper
			
			S-iPrompt~\cite{Wang2022SPromptsLW} & NeurIPS 2022 & \val{81.40}{3.3} & \val{80.80}{0.20} & \val{59.00}{6.8} & \val{57.90}{0.2} & \val{62.7}{2.2}  & \val{65.8}{0.9} & \val{64.2}{5.8}  & \val{63.4}{2.8} \\     % From DCE paper

            DCE~\cite{li2025addressing} & ICML 2025 & \val{84.6}{3.0} & \val{84.4}{0.2} & \val{64.3}{6.0} & \val{63.5}{0.5} & \val{80.1}{0.70} & \val{84.8}{0.30} & \val{74.6}{6.5} & \val{71.8}{4.2} \\      % Reported by the authors

            DualCP~\cite{Wang2025DualCPRD} & AAAI 2025 & N/A & N/A & N/A & 60.13 & N/A & 88.10 & N/A & 82.16 \\

			\midrule
            CONEC-LoRA & & \valb{86.29}{1.53} & \valb{86.43}{0.38} & \valb{66.79}{2.94} & \valb{66.42}{0.38} & \valb{88.88}{0.54} & \valb{90.24}{1.64} & \valb{88.43}{2.21} & \valb{88.21}{0.88} \\
			CONEC-LoRA with the oracle & & \val{86.98}{1.55} & \val{87.63}{0.20} & \val{68.16}{2.96} & \val{68.48}{0.22} & N/A & N/A & \val{88.98}{2.24} & \val{89.68}{0.68} \\
            
			\bottomrule
		\end{tabular}
	}
\end{table*}
\Cref{tab:DIL_accuracies} reports the accuracy of CONEC-LoRA and other consolidated algorithms. It is clearly seen that our algorithm outperforms other algorithms across all datasets with significant margins. In the CDDB case, our algorithm outperforms DualCP by $6\%$, while CONEC-LoRA surpasses DCE by $2\%$ in the office-home problem. Ours is also superior in the CORe50 problem, with an over $3\%$ margin, despite the fact that the test samples belong to 3 unseen domains rather than the 8 domains the network is trained on, which demonstrates the generalization capability of CONEC-LoRA. Last but not least, our method outperforms DCE with over $2\%$ margins in the challenging DomainNet problem. This finding clearly demonstrates the advantage of our approach based on the combination of task-shared and task-specific LoRAs, the concept of the stochastic classifier, and the introduction of an auxiliary network for domain ID predictions. On the other hand, we find that the gap between that with and without an oracle is negligible, i.e., $<1\%$. This confirms the efficacy of our auxiliary network, which produces reliable domain ID predictions for the selection and isolation of task-specific LoRA. Such a parameter isolation approach assures that task-specific knowledge can be maintained and effectively combined with shareable information to deliver accurate classifications.  

\subsection{Domain Classification Accuracies}

\begin{table*}[t]
    \caption{Domain classification accuracy}
    \label{tab:domain_classification}
    \centering
    \begin{tabular}{|l|l|ccc|}
        \toprule
         Method & Venue & CDDB-Hard & DomainNet & Office-Home  \\
         \midrule
         S-iPrompts~\cite{Wang2022SPromptsLW,Wang2025BoostingDI} & NeurIPS 2022 & $81.79$ & $80.35$ & N/A \\
         PINA-D+SOYO~\cite{Wang2025BoostingDI} & CVPR 2025 & $89.84$ & $85.37$ & N/A \\
         CONEC-LoRA & & \valb{94.70}{2.09} & \valb{88.52}{2.58} & \valb{83.72}{5.74} \\
         \bottomrule
    \end{tabular}

\end{table*}

Since our approach relies on the auxiliary network for the parameter selection strategy, \Cref{tab:domain_classification} compares the domain classification accuracy of CONEC-LoRA, SOYO~\cite{Wang2025BoostingDI}, and S-iPrompt \cite{Wang2022SPromptsLW}. Note that both SOYO and S-iPrompt utilize an external network for parameter isolation strategies. The advantage of our auxiliary network is clearly reported in \Cref{tab:domain_classification}, where it outperforms other methods with significant margins. It beats SOYO by a $5\%$ margin in the CDDB problem while surpassing SOYO with $3\%$ gap in the domainNet problem. We don't show the domain classification accuracies of other methods in the Office-Home problem because they are absent in their original papers. On the other hand, our auxiliary network performs well in the Office-Home problem, where it attains $83.72\%$ accuracy. Note that the novelty of our auxiliary network is found in at least 3 facets: 1) it applies the concept of ball-generator loss to cope with the synthetic sample-bias problem; 2) it is structured under the different-depth network structure to exploit the strength of intermediate representations; 3) it makes use of the transformation module under a fixed backbone network.  

\subsection{Ablation Studies}

\begin{table*}[t]
    \caption{Ablation study on the CDDB-Hard dataset. The columns show the average and last domain-incremental learning (DIL) accuracies for modifications to the domain-incremental classification module, as well as the domain classification accuracy (DC) for modifications to the domain classification module, across five different orders of domains.}
    \label{tab:ablation}
    \centering

    \begin{tabular}{|l|cc||l|c|}
        \toprule
         Description & DIL Average & DIL Last & Description & DC Average \\
         \midrule
\rowcolor{lightgray}         CONEC-LoRA                         & \val{88.43}{2.21} & \val{88.21}{0.88} & CONEC-LoRA & \val{94.73}{2.08} \\
         Cosine classifier instead of the stochastic classifier & \val{87.42}{3.18} & \val{87.80}{2.40} & Without the ball-generator loss                  & \val{90.53}{3.66} \\
         Linear classifier instead of the stochastic classifier & \val{81.26}{3.47} & \val{77.96}{6.74} & Without the intermediate domain classifiers & \val{94.31}{2.46} \\
         Only task-specific LoRAs                               & \val{88.79}{2.54} & \val{87.62}{3.80} & & \\
         \bottomrule
    \end{tabular}
    
\end{table*}

\Cref{tab:ablation} reports the numerical results of our ablation studies, where the contribution of different components is analyzed. It is observed that the use of the cosine classifier in lieu of the stochastic classifier reduces the domain incremental learning accuracy. A significant performance deterioration is observed when changing the classifier to the linear classifier, i.e., CONEC-LoRA suffers from over $10\%$ loss of performance. This finding confirms our claim that the stochastic classifier plays a key role in our algorithm. On the other hand, the importance of the shared LoRAs is perceived by configuring our method with only the task-specific LoRA. CONEC-LoRA suffers from performance degradation using only the task-specific LoRA without the shared LoRA, i.e., the final accuracy deteriorates, showing the CF problem due to the absence of task-invariant features.  

We also evaluate the performance of our auxiliary classifier without the ball generator loss. The removal of ball generator loss results in a decline in domain classification accuracy by about $4\%$, ultimately affecting the domain incremental learning accuracy. On the other hand, using a single classifier instead of multiple local classifiers, i.e., the different-depth network structure, reduces domain classification accuracy. The different-depth network structure fully exploits the intermediate representations in identifying the correct domain labels.   

\subsection{Robustness Analysis}

\begin{table}[t]
    \caption{Robustness of CONEC-LoRA to the variations of $\lambda_1$ coefficient on the domain-incremental learning (DIL) accuracy and $\lambda_2$ coefficient on the domain classification (DC) accuracy on the CDDB-Hard dataset.}
    \label{tab:robustness_analysis}
    \centering

    \begin{tabular}{|l|cc||l|c|}
        \toprule
         $\lambda_1$ & DIL Average & DIL Last &  $\lambda_2$ &  DC Average\\
         \midrule
                     $3$ & \val{88.36}{2.24} & \val{88.23}{1.12} & $1$   & \val{94.23}{2.32} \\
                     $4$ & \val{88.43}{2.21} & \val{88.23}{0.94} & $1.5$ & \val{94.66}{2.09} \\
\rowcolor{lightgray} $5$ & \val{88.43}{2.21} & \val{88.21}{0.88} & $2$   & \val{94.73}{2.08} \\     % Our default setting
                     $6$ & \val{88.49}{2.24} & \val{88.20}{1.03} & $2.5$ & \val{94.83}{2.05} \\
                     $7$ & \val{88.57}{2.22} & \val{88.34}{1.08} & $3$   & \val{94.96}{2.00} \\
         \bottomrule
    \end{tabular}
    
\end{table}

We evaluate the robustness of the CONEC-LoRA to the variations of $\lambda_1$ and $\lambda_2$, which are crucial coefficients to control the contributions of loss functions. \Cref{tab:robustness_analysis} shows the robustness of the proposed method to the variations of each coefficient while keeping the other fixed. It is seen that CONEC-LoRA is not sensitive to different trade-off constants $\lambda_1,\lambda_2$ where variations of these parameters do not significantly affect the performances of CONEC-LoRA. This finding confirms the robustness of CONEC-LoRA against different trade-off constants steering the contributions of loss functions. For the sake of convenience, the parameters are simply set to $\lambda_1=5,\lambda_2=2$ for our experiments. 

\subsection{UMAP Analysis}

\begin{figure}[t]
    \centering
    \captionsetup[subfigure]{justification=centering}

    \begin{subfigure}
        \centering
        \includegraphics[width=0.45\linewidth]{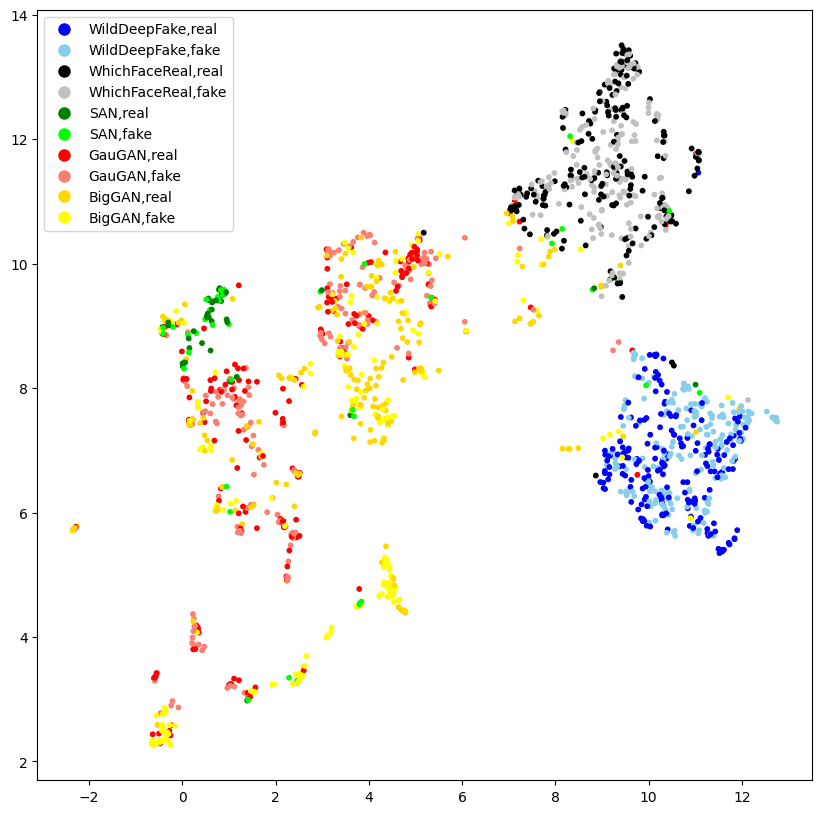}
    \end{subfigure}
    \hfill % \vspace{0.5em}
    \begin{subfigure}
        \centering
        \includegraphics[width=0.45\linewidth]{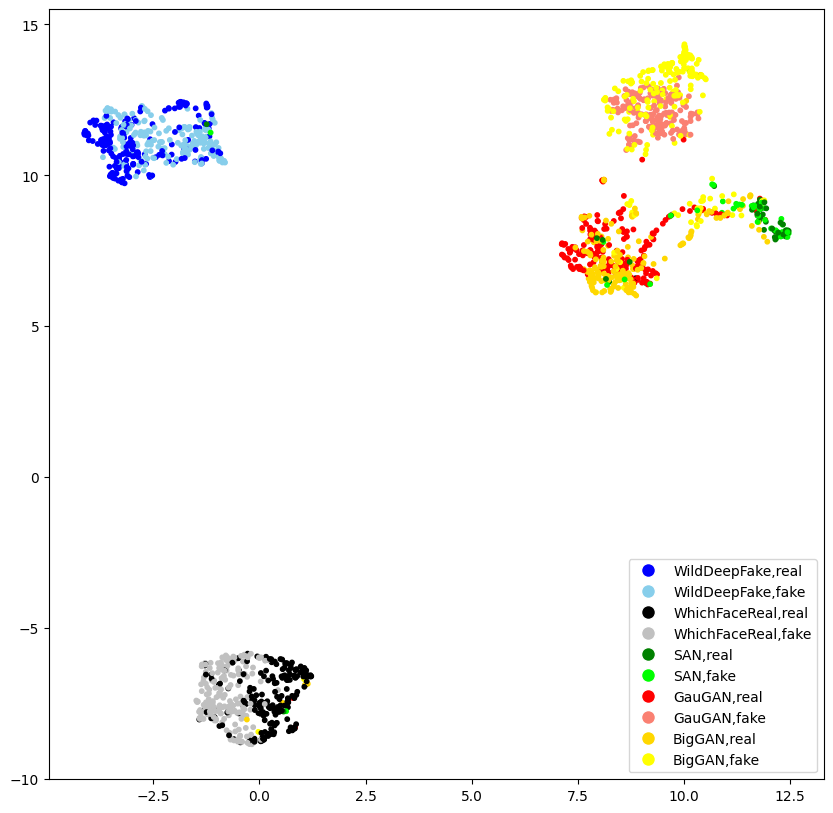}
    \end{subfigure}

    \caption{UMAP plots on the embedding space of the model before training on the left and after training on the right on five domains of the CDDB-Hard dataset.}
    \label{fig:UMAP}
\end{figure}

\cref{fig:UMAP} exhibits the UMAP~\cite{mcinnes2018umap} graphs of the network on 5 domains of the CDDB dataset before training and after training with CONEC-LoRA. It is clearly seen that before the training process, the embeddings are scattered in the feature space, and the pre-trained model seems to be confused in distinguishing samples of different domains and classes. After the training process, CONEC-LORA exhibits strong discriminative features, where samples of different domains are projected into different locations of the latent spaces. Besides, our algorithm is capable of separating different classes within the domains. This finding also confirms that our approach is relatively robust to the CF problem, where our model recognizes all sequentially presented domains well, i.e., the UMAP plot refers to the model after training on the last domain.  

\section{Conclusion}
This paper proposes continual knowledge consolidation low rank adaptation (CONEC-LoRA) as a solution to domain incremental learning (DIL) problems. CONEC-LoRA resolves the three shortcomings of existing methods: parameter isolation, inaccurate selection of task-specific parameters, and the use of traditional classifiers. CONEC-LoRA features a synergy between shared LoRA and task-specific LoRA while putting forward the auxiliary domain classifier, which includes the ball-generator loss, the different-depth network structure, and the transformation module. It integrates the concept of stochastic classifiers having class mean and variance vectors drawn from a distribution. The efficacy of CONEC-LoRA has been rigorously evaluated with the four benchmark problems of DIL, where it beats recently published works across all problems with over $5\%$ margins. Its domain classification accuracies are also superior to existing works, while the ablation study substantiates the advantage of each learning component. The robustness analysis confirms that our algorithms are not sensitive to variations of loss coefficients. Our future work is devoted to studying the data scarcity problem in DIL.  

\bibliographystyle{IEEEtran}
\bibliography{references}

\end{document}